\documentclass[letterpaper, 10 pt, conference]{ieeeconf}  

\IEEEoverridecommandlockouts                              

\usepackage{graphics} 
\usepackage{epsfig} 
\usepackage{mathptmx} 
\usepackage{times} 
\usepackage{amsmath} 
\usepackage{amssymb}  
\usepackage{caption}
\usepackage{subcaption}
\usepackage{xcolor}
\usepackage{multirow}
\usepackage{booktabs}
\usepackage{adjustbox}
\let\labelindent\relax
\usepackage{enumitem}
\DeclareGraphicsExtensions{.pdf, .PDF, .png, PNG, .jpeg, .JPG}

\title{\LARGE \bf
Infrastructure-based End-to-End Learning and \\ Prevention of Driver Failure}

\author{Noam Buckman$^{1*}$, Shiva Sreeram$^{1,2*}$,  Mathias Lechner$^1$, Yutong Ban$^1$,\\  Ramin Hasani$^1$, Sertac Karaman$^3$, Daniela Rus$^1$
\thanks{*equal contribution}%
\thanks{This work is supported by the Toyota Research Institute (TRI). This article solely reflects the opinions and conclusions of its authors and not TRI or any other Toyota entity. Their support is gratefully acknowledged.}
\thanks{$^{1}$Computer Science and Artificial Intelligence Laboratory, Massachusetts Institute of Technology, Cambridge, MA 02139, USA 
{\tt\small [nbuckman, sasreera, mlechner, yban, rhasani, rus] at mit.edu}}%
\thanks{$^{2}$California Institute of Technology, Pasadena, CA 91125, USA }%
\thanks{$^{3}$Laboratory of Information and Decision Systems, Massachusetts Institute of Technology, Cambridge, MA 02139, USA
{\tt\small sertac@mit.edu}}%
}

\begin{document}

\maketitle
\thispagestyle{empty}
\pagestyle{empty}

\begin{abstract}

Intelligent intersection managers can improve safety by detecting dangerous drivers or failure modes in autonomous vehicles, warning oncoming vehicles as they approach an intersection. In this work, we present FailureNet, a recurrent neural network trained end-to-end on trajectories of both nominal and reckless drivers in a scaled miniature city. FailureNet observes the poses of vehicles as they approach an intersection and detects whether a failure is present in the autonomy stack, warning cross-traffic of potentially dangerous drivers. FailureNet can accurately identify control failures, upstream perception errors, and speeding drivers, distinguishing them from nominal driving. The network is trained and deployed with autonomous vehicles in the MiniCity. Compared to speed or frequency-based predictors, FailureNet's recurrent neural network structure provides improved predictive power, yielding upwards of 84\% accuracy when deployed on hardware. 

\end{abstract}

\section{INTRODUCTION}
Safety is critical for the adoption of autonomous vehicles (AVs) on roads, especially as an increasing number of vehicles are deployed on the road. Given that failures and errors will always exist, methods must be developed for identifying issues with autonomous vehicles and alerting vehicles with enough time to take action. Infrastructure-based methods, such as intelligent intersection managers, can observe drivers for longer duration for improved failure detection. In addition, control of the intersection provides an extra level of safety, especially for cross-traffic collisions. 

In this work, we consider an intelligent traffic light that monitors vehicles for failures and warns oncoming traffic to prevent collisions. Existing approaches such as driver monitoring systems require in-cabin sensor placement for driver monitoring which can capture more information but requires access to the vehicle itself, whereas an external monitor does not require access to the vehicle. Furthermore, current approaches are typically limited to vehicles observed within the field-of-view of the vehicle, whereas external monitoring from an intersection manager can monitor vehicles as they approach an intersection. 

In our approach, an intersection manager observes a vehicle's trajectory as it drives near an intersection and uses FailureNet, a recurrent neural network (RNN), to detect whether a driver's behavior is caused by a planning or actuator failure (Fig.~\ref{fig:fig1}). Our learning-based approach is trained to detect failures from generated data within the MiniCity, a 1/10th miniature city  where multiple autonomous vehicles are deployed simultaneously. We induce vehicle failures in the scaled hardware, ranging from control failures (injecting noise to speed and steering) to perception failure, and train FailureNet on this novel dataset. We demonstrate the accuracy of FailureNet and our ability to warn oncoming traffic by deploying in the MiniCity with multiple vehicles and compare to multiple baseline approaches. 

In summary, we make the following contributions:

\begin{figure}
    \centering
    \begin{subfigure}{0.49\columnwidth}
        \includegraphics[width=0.99\columnwidth]{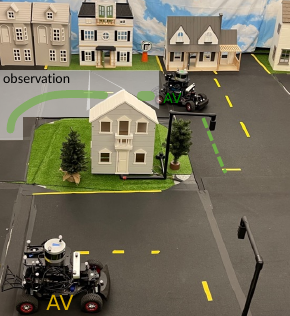}
    \caption{Nominal driver}\label{fig:nominal_1}
    \end{subfigure}
    \begin{subfigure}{0.49\columnwidth}
        \includegraphics[width=0.99\columnwidth]{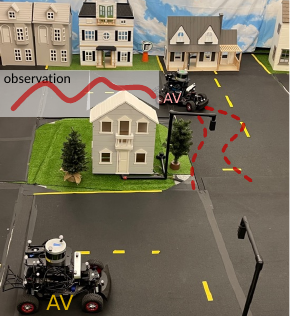}
    \caption{Reckless driver}\label{fig:reckless_1}
    \end{subfigure}
    \caption{An intelligent intersection warns oncoming traffic of dangerous drivers by observing driving behaviors as vehicles approach the intersection}
    \label{fig:fig1}
\end{figure}

\begin{enumerate}[leftmargin=*]
    \item An end-to-end algorithm for detecting vehicle failures and warning oncoming drivers using intelligent traffic lights;
    \item A pipeline for generating and deploying failure-induced driving styles for training various failure detectors; 
    \item Training and evaluation of FailureNet in a physical 1/10th-scaled MiniCity with fully autonomous vehicles and intelligent traffic infrastructure.
\end{enumerate}

\section{RELATED WORKS}

\subsection{Monitoring Ego Driver}
The ability to detect failures in AV stacks or anomalies in human drivers is crucial for trust in AVs. Recent work \cite{Hecker2018,Svegliato2019} has explored methods for introspective monitoring of the AV stack for faults and anomalies by observing the state of the vehicle. 
For human drivers, neural networks learn from on-board vehicle diagnostics to identify driver anomalies~\cite{Zhang2017a} and ~\cite{Johnson2011b,Vasconcelos2017} use onboard cellphone data to train a network to identify different driving styles.
\cite{Quintero2010} use a simulator to generate erratic driving and detect anomalies with a neural network. Other learning-based approaches use supervised learning~\cite{Siddiqui2016uai} or reinforcement learning~\cite{Wu2021arxiv} to detect rare events in time-series data. 
In \cite{Ryan2021}, a Gaussian Processes models nominal human driver based on pre-recorded human driver trajectories, and identify anomalies in an AV if observed steering is outside a 95\% confidence interval.
Non-learning approaches include identifying faults with a Kalman filter~\cite{Kawashima2003, Duan2005} and analyzing the frequencies in driver steering~\cite{Takei2005} to identify driver fatigue in a simulation.
In all these methods, the network requires access to the vehicle's internal state, from driver inputs to software outputs, to accurately identify driver anomalies which limits monitoring.

\subsection{Monitoring Surrounding Traffic}
For autonomous perception and planning, many systems monitor surrounding vehicles to predict the driver's state or agent's future motion.
In \cite{DiBiase2021}, a dataset of anomalies is generated, and a detector is trained on images to identify anomalous scenes. In \cite{Doshi2009}, the eyesight of other drivers is used to predict lane change intent. \cite{Fletcher2009} use eye-gaze observations to predict inattention for collision avoidance. In~\cite{wiederer2022}, a density function is learned from simulated abnormal and normal trajectories to calculate low-density, anomalies. In contrast, we directly predict failure modes generated by real hardware and software failures.

Instead of predicting the driver's state or driving behavior, trajectory predictors predict future trajectories directly. \cite{Morton2017} use an LSTM to predict acceleration profiles and compare to classic driver models such as the intelligent driver model (IDM). They evaluate on the NGSIM Highway dataset comparing predicting vehicle position and actual position.   
\cite{Salzmann2020} uses a graph-based LSTM to predict dynamically feasible trajectories for robots navigating around multiple agents. In both examples, robots and agents act nominally without failures present. In addition, predicting entire trajectories during rare failures may not be necessary or possible, especially without explicitly modeling whether a failure is occurring.

\subsection{Infrastructure-based Systems}
Intelligent intersection managers can be used to both observe traffic participants and direct drivers to prevent collisions. \cite{Shirazi2017,Bjorklund2005} discuss various approaches for monitoring intersections.
In \cite{Sun2022}, multi-camera views are fused to predict incoming traffic for an intersection.
\cite{Phillips2017} uses an LSTM to predict driver intention at intersections. In both, datasets that typically only experience nominal driving behavior are used, and rarely capture dangerous driving behaviors.
In~\cite{Lefevre2012}, a deep Bayesian network is used to predict driver intentions at intersections and validated with field experiments. In~\cite{raja2022ai}, a V2X anomaly framework is evaluated on simulated trajectories to identifier outliers, whereas in this work we consider failure-induced anomalies in physical hardware.

\begin{figure}[t]
    \centering
    \includegraphics[width=0.8\columnwidth,trim={0 0 200px 50px}, clip]{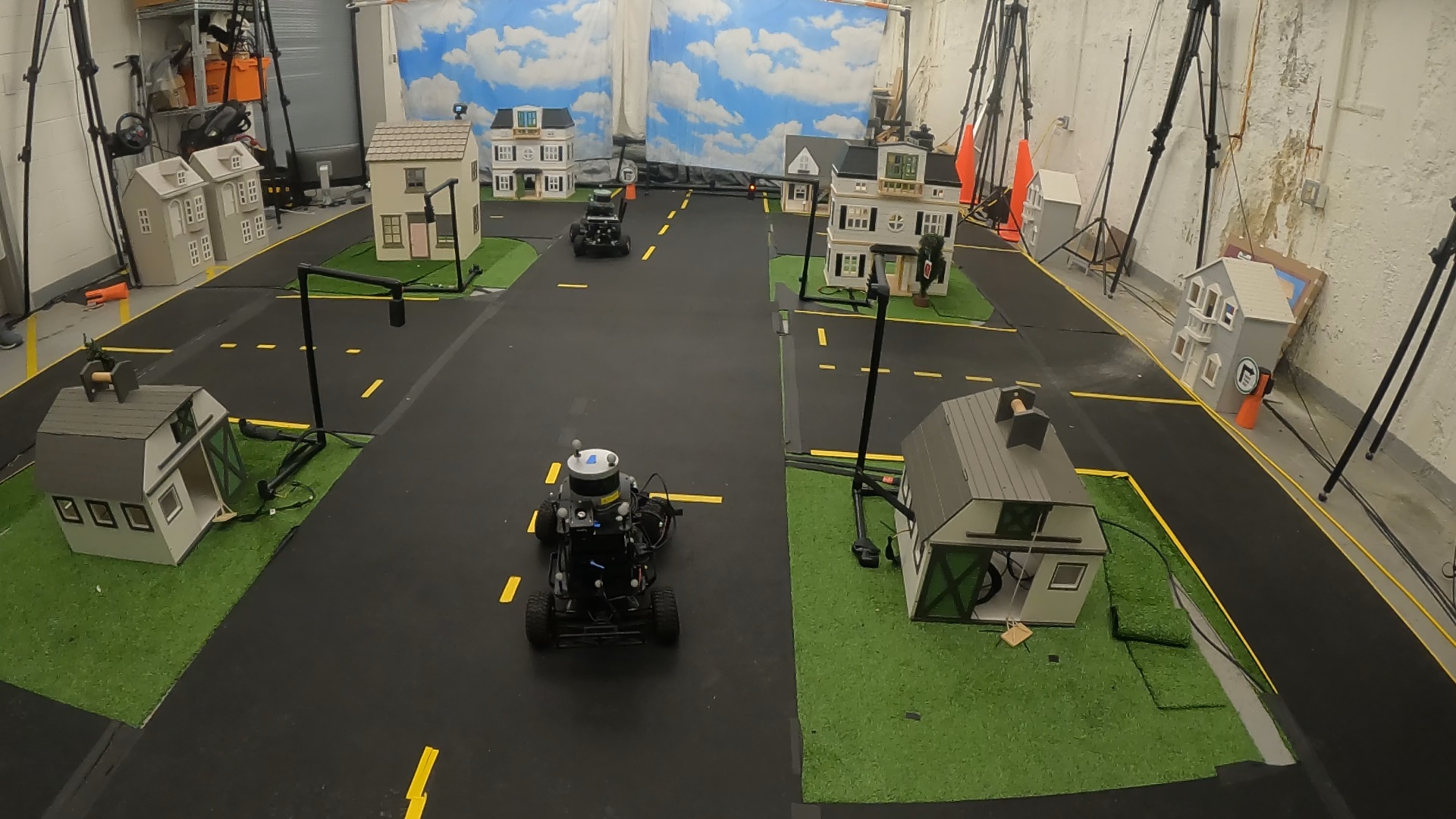}
    \caption{Vehicle Driving Through the MiniCity Intersection}
    \label{fig:minicity}
\end{figure}

Once a dangerous driver is detected, an intelligent intersection manager should actively warn oncoming traffic of dangers. 
\cite{Salim2007} use a simulator to validate a collision detection algorithm for cross traffic at intersections.
For preventing collisions, \cite{Kowshik2011a} propose a hybrid scheduler-controller to provide provably safe intersections and in ~\cite{Dresner2008a}, a supervised intelligent reservation manager modifies existing reservations in the presence of catastrophic failures.
In ~\cite{Yu2019, Feng2018}, full-scale cars are deployed on closed courses to evaluate human driver acceptance of V2I recommendations.  However, given the inherent dangers with full-scale testing of failure modes at intersections, previous work have been deployed either purely in simulation~\cite{Dresner2008a} or deployed with nominal driving behaviors~\cite{Yu2019}. In this work, we deploy on real hardware in a miniaturized city (Fig.~\ref{fig:minicity}) with multiple autonomous vehicles of various driving behaviors. This allows us to train and deploy reckless drivers using a physical platform without similar safety concerns.

\section{Problem Statement}\label{sec:problem_statement}
The goal of this work is to successfully identify vehicles with planning or sensor failures before the vehicles arrive at an intersection. 
Specifically, we consider an intersection and the surrounding roads flowing into the intersection that is monitored by an intelligent intersection observer and manager. We assume that under normal operation (nominal driving), each vehicle $j$ navigates to various locations in the city autonomously, with a high-level route planning, low-level path planning, and motion control. A vehicle failure is defined as a significant degradation of one or more sub-components of the autonomous vehicle, for example, decision-making, perception, or low-level control. We assume that a vehicle's failure persists through the duration of driving and is represented by a latent failure variable, $z_j \in \{0, 1\}$ where $1 = Unsafe$, $0 = Safe$. 

The goal of the intersection manager is to observe the vehicles and (1) detect whether a vehicle is failing and if so, (2) mitigate intersection collisions by warning oncoming traffic.  The intersection manager only has access to information observable externally. Specifically, the intersection manager observes pose of each of agent $p_{j,t} = [x_t, y_t, \theta_t]$  and the goal of traffic light is to provide an estimate $\hat{z}_{j,t}$ of whether vehicle $j$ is experiencing a failure, and if so, communicate to incoming vehicles.

\subsection{Failure Modes}\label{sec:failure_modes}
On-road collisions can occur due to various types of vehicle failures. In this work, we focus on identifying failure modes that manifest in the driving behavior of the vehicle before the point of collision at an intersection. 
While some failures may present when it is too late to mitigate a collision, we focus on driving or control failures that may manifest as the vehicle approaches the intersection. 
We consider four types of vehicle failures of reckless driver profiles for the vehicles, motivated by reckless human driving and autonomous vehicle failures.

\subsubsection{Random Periodic Control Failure}
The first type of failure is a random additive noise applied to the control output of the vehicle: steering and velocity. This failure mode is chosen to demonstrate a persistent random vehicle failure or poor driver abilities.  Specifically, a random steering and speed noise is added to the desired steering and speed outputted by the autonomy stack.

\begin{equation}
    \delta_t = \delta_{command,t} + \epsilon_{\delta,t},~~~v_t = v_{command,t} + \epsilon_{v,t},
\end{equation}
where $v_t$ and $\delta_t$ are the instantaneous velocity and steering, $v_{t,command}, \delta_{t, command}$ are the commanded velocity and steering by the autonomous controller, and $\epsilon_{v}, \epsilon_{\delta}$ are random noise variables. The noise is added to the control output (steering, speed) and not the actuator output (motor current).

The velocity and steering profiles are chosen as
\begin{equation}
    \epsilon_{\delta,t} = A_\delta \sin \Big( \frac{2\pi}{T_\delta} t \Big)
\end{equation}
and
\begin{align*}
    & \epsilon_{v,t} = A_{v,T} \cdot Hold(T_{v}) 
    &A_{v,T}  \sim \mathcal{U}_{[a,b]}
\end{align*}
where $A_{\delta}$ is the magnitude of steering noise, $\mathcal{U}_{[a,b]}$ is a uniform distribution with lower and upper limits $a$ and $b$, respectively, and $Hold(T_{v})$ maintains a constant value for duration $T_v$. We choose time constant $T_\delta$ and $T_v$ such that the noise propagates to meaningful periodic movement of the vehicle.

\begin{figure}[t]
    \begin{subfigure}[c]{0.47\columnwidth}
        \includegraphics[width=0.99\columnwidth]{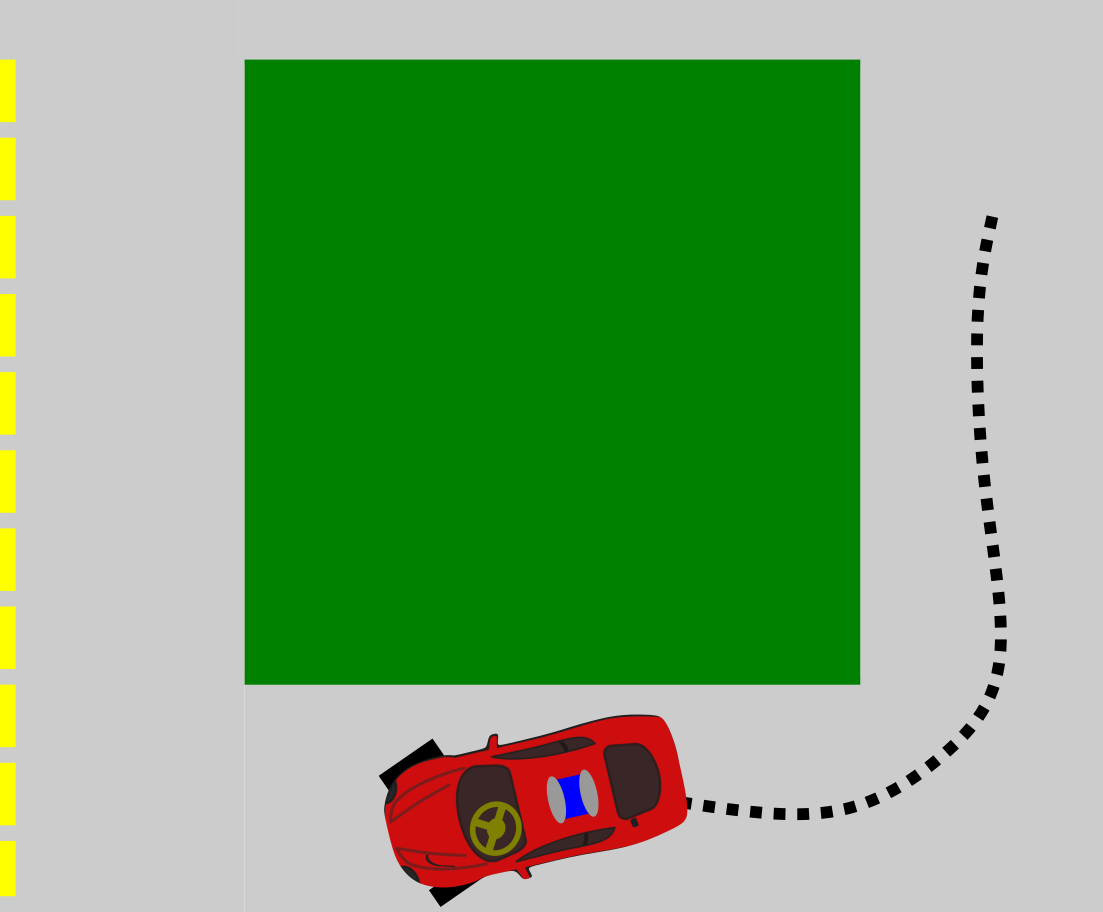}
        \caption{Periodic Steering and Speed}
        \label{fig:drawing_periodic}
    \end{subfigure}\hfill
    \begin{subfigure}[c]{0.47\columnwidth}
        \includegraphics[width=0.99\columnwidth]{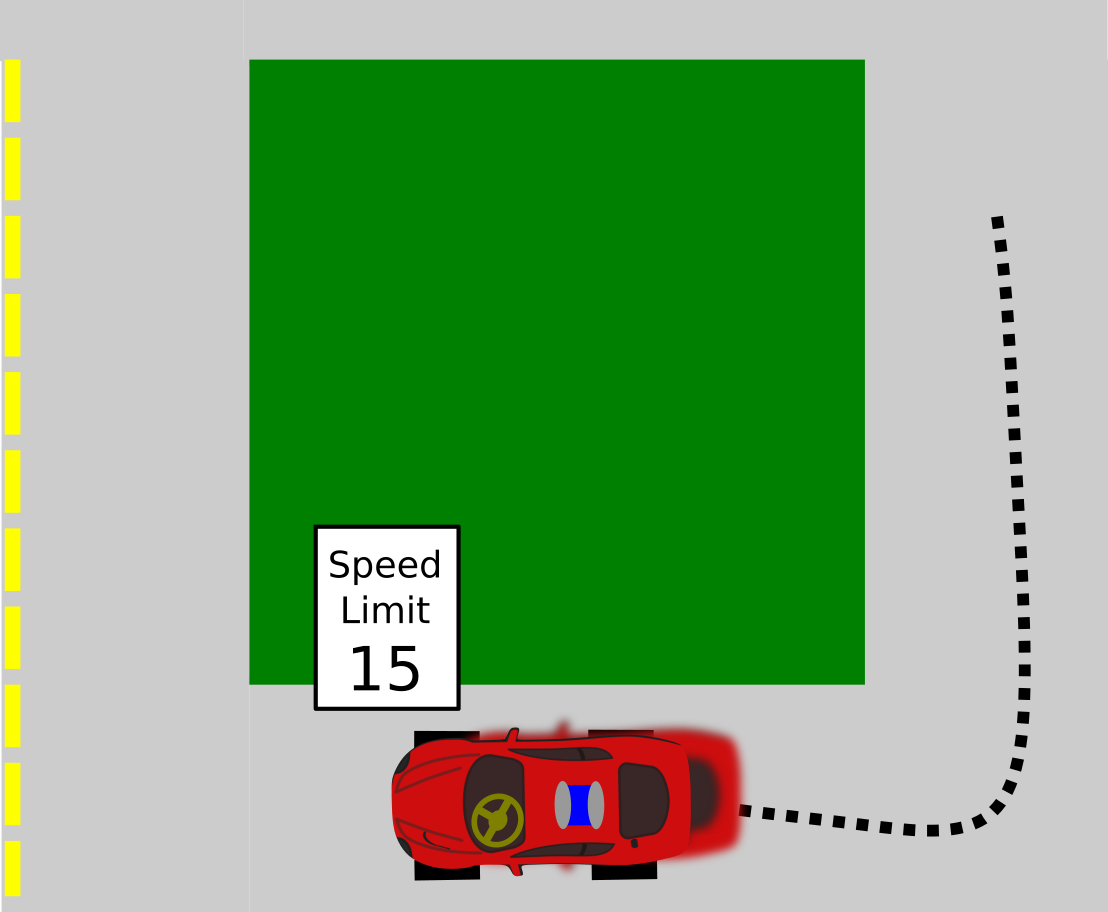}
        \caption{Speeding Driver}
        \label{fig:drawing_speeding}
    \end{subfigure}       
    \vskip\baselineskip
    \begin{subfigure}[c]{0.47\columnwidth}
        \includegraphics[width=0.99\columnwidth]{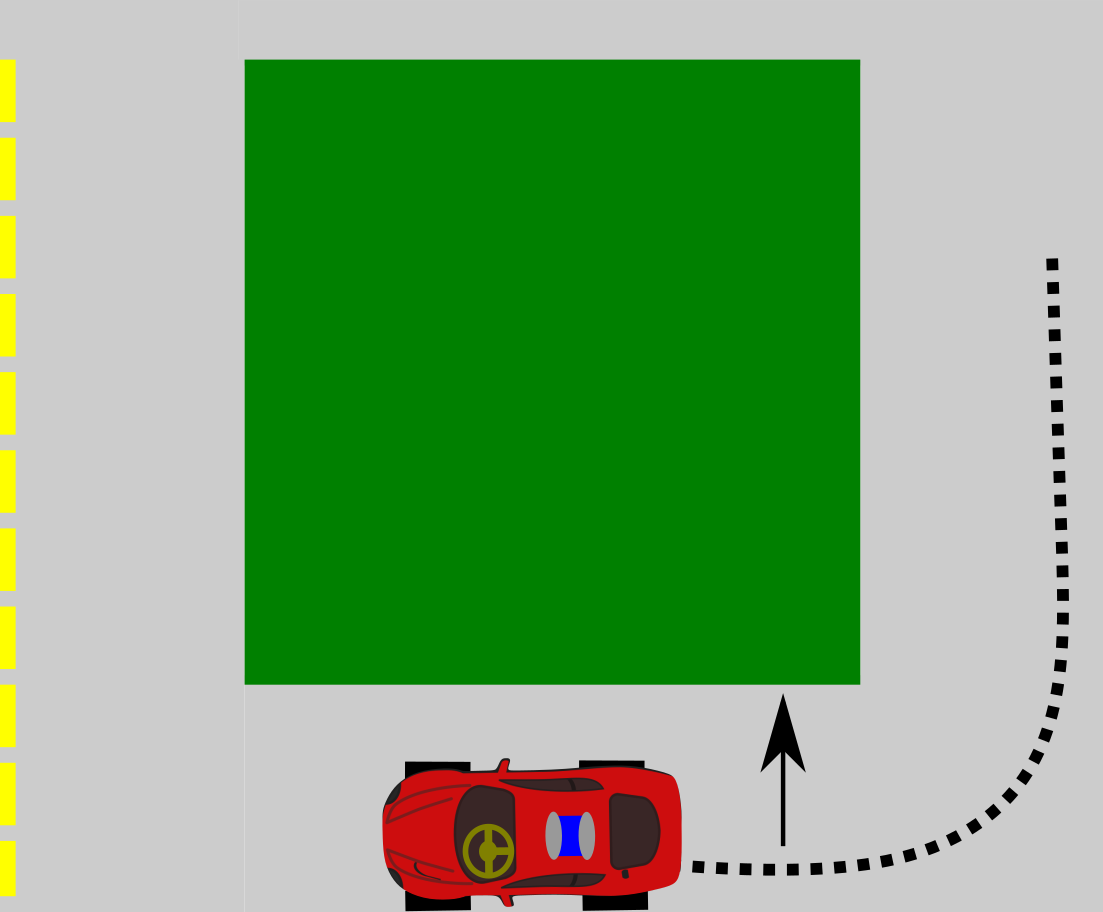}
        \caption{Lane Detection Offset}
        \label{fig:drawing_lane_detection}
    \end{subfigure}\hfill    
    \begin{subfigure}[c]{0.47\columnwidth}
        \includegraphics[width=0.99\columnwidth]{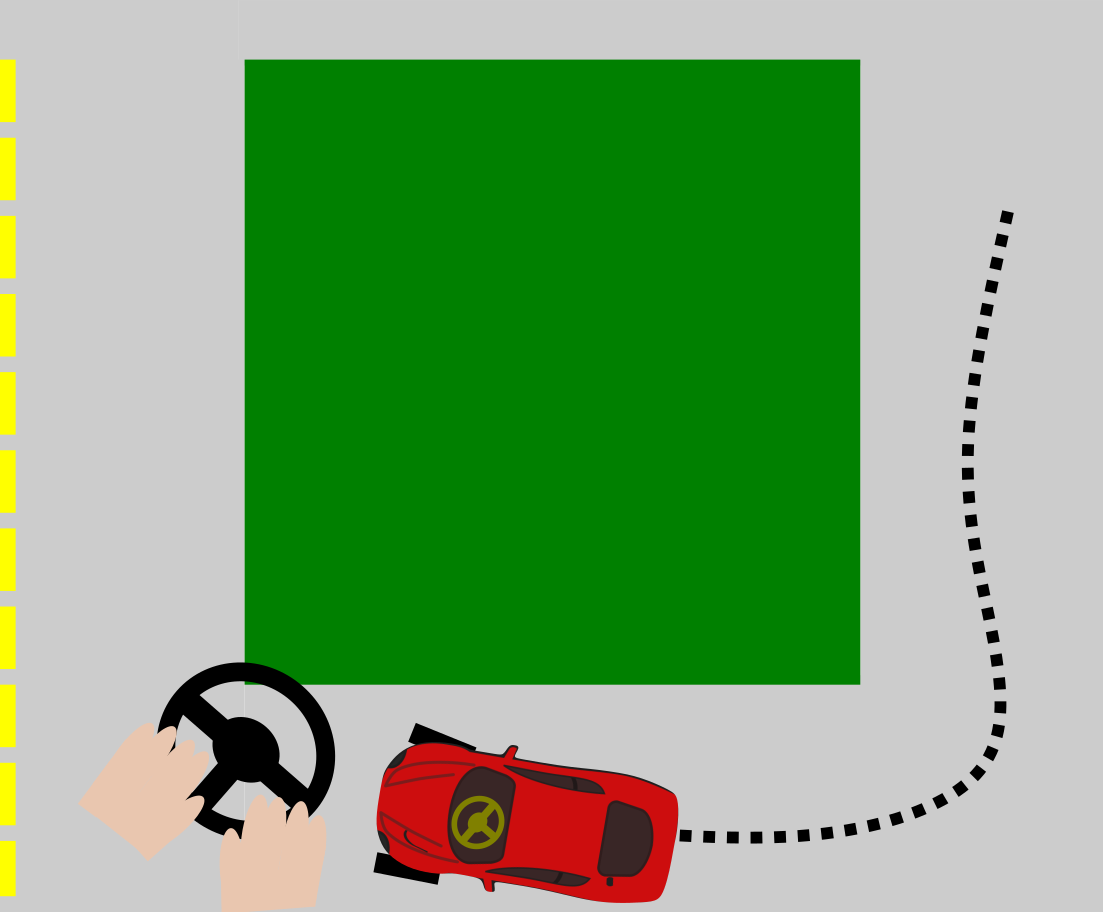}
        \caption{Manual Reckless Driver}
        \label{fig:drawing_manual}
    \end{subfigure}        
    \caption{Failure Modes. Four failure modes are deployed into the MiniCity to capture varying types of anomalies, ranging from low-level control errors (a), (d) to high-level planning failures (b)(c).}
    \label{fig:hardware_failures}
\end{figure}
\subsubsection{Lane Detection Offset Failure}

Upstream failures in the perception of the vehicle, such as a failing lane detector, can lead to observable and dangerous scenarios. We consider a failing lane detector that outputs an incorrect lane line. Similar to the other failure modes, we consider a non-catastrophic failure such as a biased lateral shift of the outer lines. For each outer lane line detected, the line is shifted laterally by a distance $\bar{s}$. The shifted lane line is then processed by the vehicle's path planner which generates a centerline that is shifted by approximately $\bar{s}/2$.

\subsubsection{Speeding Driver}
Speeding drivers were a contributing factor in 29\% of all deaths on the road totaling 11,258 fatalities~\cite{NationalCenterforStatisticsandAnalysis2022}.  
We simulate a speeding driver by increasing the desired speed of the driver from 0.3m/s to 0.5m/s. The steering of the vehicle is unaffected; however, the vehicle attempts to maintain a desired speed of 0.5m/s. The high speed of the vehicle leads to increased steering oscillations due to dynamic instabilities, in addition to overshooting tight turn radius in the approach to the intersection.

\subsubsection{Manual Reckless Driver}
Finally, to capture a wide breadth of driving styles, we consider a hybrid failure mode generate by allowing a human driver to command the vehicle in a reckless manner.

\begin{figure*}[t]
    \centering
    \includegraphics[width=0.85\textwidth]{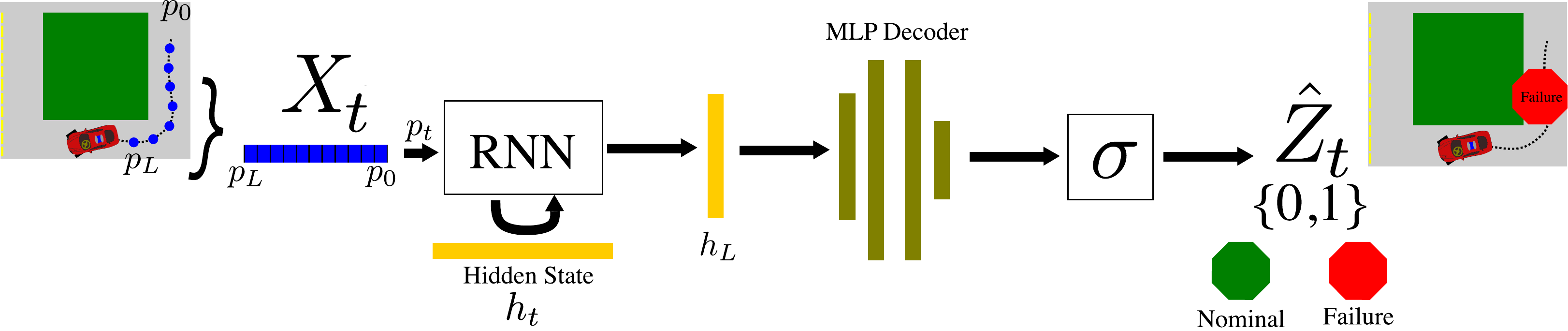}
    \caption{FailureNet Model Architecture}
    \label{fig:model_architecture}
\end{figure*}
\section{FailureNet}
We propose a learning-based approach, FailureNet, which relies purely on external pose information of each vehicle for detecting vehicle failures. In this section, we describe FailureNet's network architecture, and in the subsequent section, our training pipeline using the MiniCity.

\subsection{Model Architecture}
Our goal is to learn a function approximator \begin{equation}
    \hat{z}_t = F(X_t ; \theta)
\end{equation}
where $\hat{z}_t$ is the predicted state of the vehicle and $X_t$ is the sequence of $L$ poses starting from $p_t$ to $p_{t-L}$.  In general, individual approximators may be deployed for each vehicle $j$, $\hat{z}_{j,t}$ however, for the purpose of this work, we consider estimating the status of a single vehicle and drop $j$ for simplicity.

An end-to-end autoregressive modeling framework (Fig.~\ref{fig:model_architecture}) can be deployed to learn a proper representation from spatio-temporal input observations. To do this, we parameterize a recurrent neural network (RNN) with the following states update rule:

\begin{equation}
    h_t = g_{RNN}(p_t, h_{t-1})
\end{equation}
where $h_t \in R^{n_h}$ is the hidden state of dimension $n_h$, and $g_{RNN}$ is the non-linear recurrent cells of the model. 

We decode the hidden state through an encoder-decoder architecture, where the hidden state of the RNN compartment at the end of the input sequence, $h_T$, is decoded to output predictions via a multi-layer perceptron $f(\cdot)$, as follows:
\begin{equation}
    d = f(h_T)
\end{equation}

\noindent The decoded hidden state is then passed through a sigmoidal output layer:
\begin{equation}
    \hat{z}_t = \sigma(d_t)
\end{equation}
where $\sigma(\cdot)$ is a logistic sigmoid function and $\hat{z}_t \in (0, 1)$ corresponding to $0 = Safe$, $1 = Unsafe$.

During training, we utilize a binary cross-entropy (BCE) loss function constructed as follows:

\begin{equation}
    \mathcal{L}(z_t, \hat{z}_t) = z_t \log(\hat{z}_t) + (1-z_t) \log(1-\hat{z}_t)
\end{equation}
where $z_t$ are the ground truth labels for whether a planning or actuation failure is occurring and $\hat{z}_t$ are the predicted state. 

\subsection{Choice of the Recurrent Neural Networks}
To encode input sequences, we can use gated recurrent neural networks such as the long short-term memory (LSTMs)~\cite{hochreiter1997long} or gated recurrent units (GRUs)~\cite{Chung2014}. Moreover, recent advances in end-to-end sequence modeling frameworks in robotics environments \cite{lechner2019designing,lechner2020neural,vorbach2021causal} showed the intriguing representation learning capabilities of a new class of continuous-time neural networks called liquid time-constant networks (LTCs) \cite{Hasani2021liquid}. LTCs are nonlinear state-space models \cite{gu2022efficiently} that are described by ordinary differential equations (ODEs) \cite{chen2018neural} or in closed-form \cite{hasani2021closed} and are reduced to dynamic causal models \cite{friston2003dynamic}, a framework through which the system can learn the cause-and-effect of a given task \cite{vorbach2021causal}. 

We use the closed-form representation of liquid neural networks, named a closed-form continuous-time neural network (CfC), as a baseline in our work to equip FailureNet with the state-of-the-art sequence modeling pipeline. CfC cells are given by the following representation \cite{hasani2021closed}:

\begin{align*}
    h_t = &\sigma(-f(h_{t-1},p_t;\theta_f)t) \odot g_1(h_{t-1},p_t;\theta_{g_1}) + \\ &[1-\sigma(-[f(h_{t-1},p_t;\theta_f)]t)]\odot g_2(h_{t-1},p_t;\theta_{g_2}).
\end{align*}
Here, $f$, $g_1$, and $g_2$ are three neural network heads with a shared backbone, parameterized by $\theta_f, \theta_{g_1},$ and $\theta_{g_2}$, respectively. $p_t$ is the vehicle position, $t$ stands for input time-stamps, and $\odot$ is the Hadamard product.

\begin{figure}
    \centering
    \includegraphics[width=0.9\columnwidth,trim={400px 200px 400px 50px}, clip]{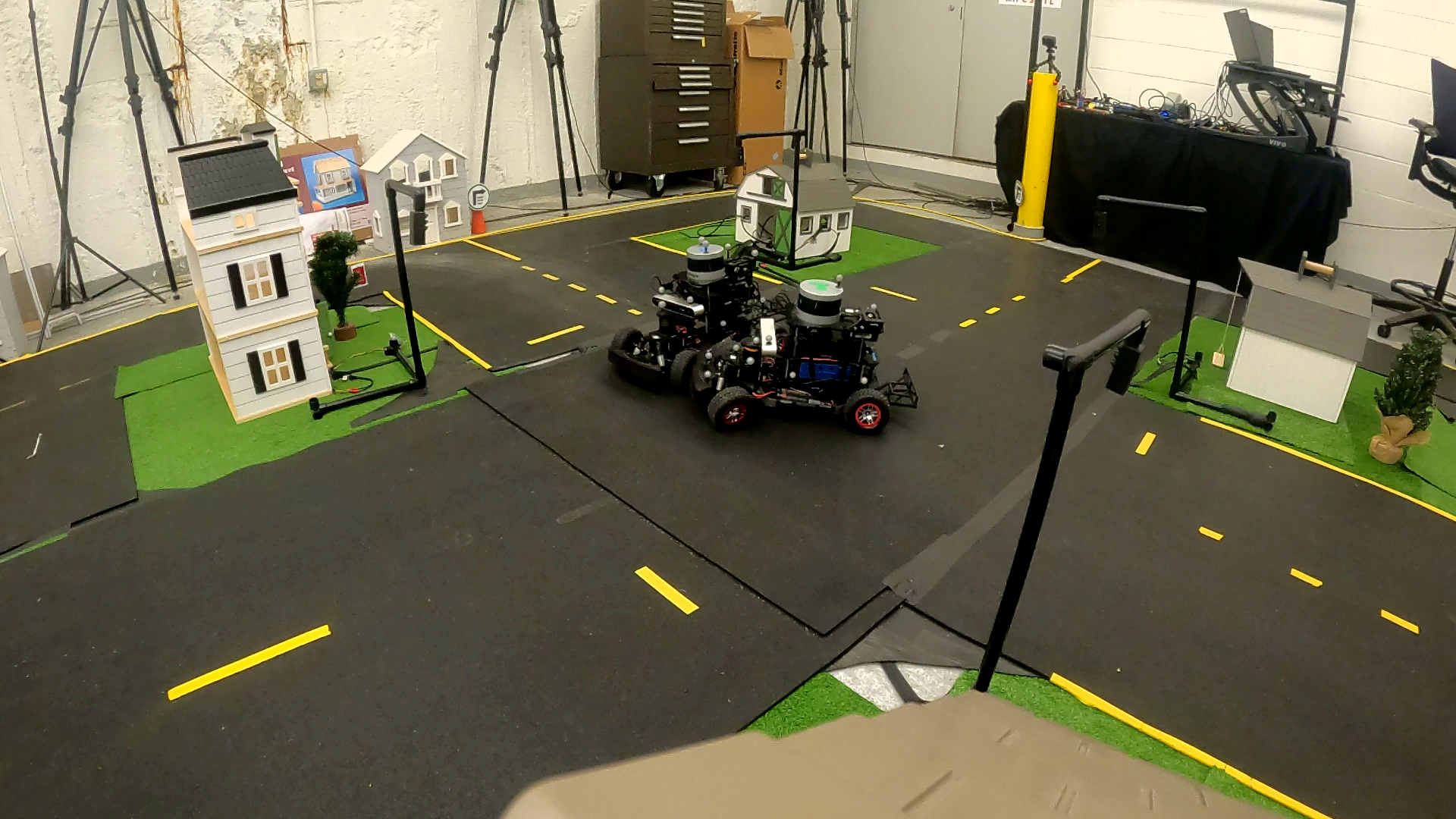}
    \caption{Collision between an AV with planning failures and a nominal autonomous vehicle inside a MiniCity intersection.}
    \label{fig:collision}
\end{figure}

\section{Training and Deploying in the MiniCity}

End-to-end learning approaches for failure detection depend heavily on ground truth labeled data. For driving environments, existing datasets typically lack failure cases. Generating these failure cases is far too dangerous with full-size vehicles. We utilize a novel scaled testing environment, the MiniCity~\cite{Buckman2022sensors}, which enables us to generate failure modes for training auto-regressive neural networks.

\subsection{The MiniCity Evaluation Platform}
The MiniCity is a 1/10th scale experimentation platform for testing and evaluating robotics research in autonomous vehicles. Scaled houses, roads, grass, and traffic lights make up a realistic aesthetic of the MiniCity, with intersections and roundabouts for simulating dangerous and interactive driving scenarios.  Each vehicle in the MiniCity consists of state-of-the-art sensors, such as a Velodyne Lidar and Zed camera, and runs a full autonomy stack from high-level mission planning to low-level control. This allows us to deactivate various components of the autonomy stack to simulate catastrophic failure and measure the impact on vehicle driving.

An external motion capture provides ground truth position for each vehicle and simulates GPS for onboard state estimation. Individual vehicles fuse multiple sensor modalities, including simulated GPS, to localize in the MiniCity, while we utilize the high-rate motion capture for collecting training data and evaluating the performance of FailureNet. In addition, a high-definition road map is provided to each vehicle for navigating in the MiniCity.

\subsection{Training on Reckless Drivers in the MiniCity}
Each vehicle in the MiniCity runs a full autonomy stack, implemented in ROS, to navigate within the city setting.
Reckless driving is simulated by injecting various failure modes in the AV stack, as described in Sec.~\ref{sec:problem_statement}. 
For high-level failure modes such as lane detection and speeding, we modify the upstream planning nodes, whereas for low-level failures such as noisy controls, we create a noisy driver ROS node that injects random noise at the output. 
Each intersection contains a signalized traffic light that communicates with vehicles over ROS. Figure ~\ref{fig:collision} shows an example failing AV colliding with a cross-traffic driver.

\begin{figure}
\begin{subfigure}[b]{0.49\columnwidth}
    \centering
    \includegraphics[width=0.99\columnwidth,trim={5px, 8px, 0, 0},clip]{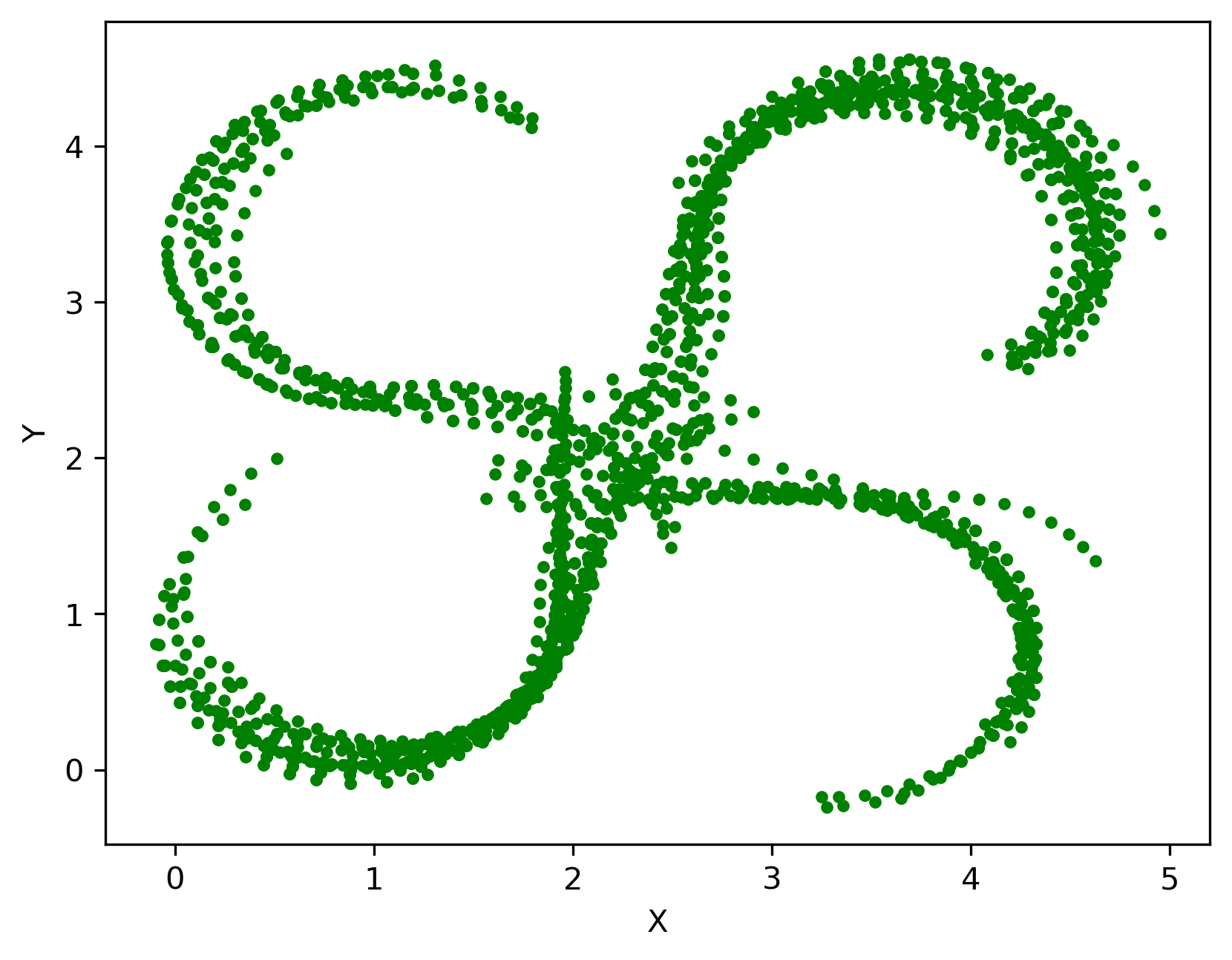}
    \caption{Periodic Noise}
    \label{fig:Noise1}
\end{subfigure}
\begin{subfigure}[b]{0.49\columnwidth}
    \centering
    \includegraphics[width=0.99\columnwidth,trim={5px, 8px, 0, 0},clip]{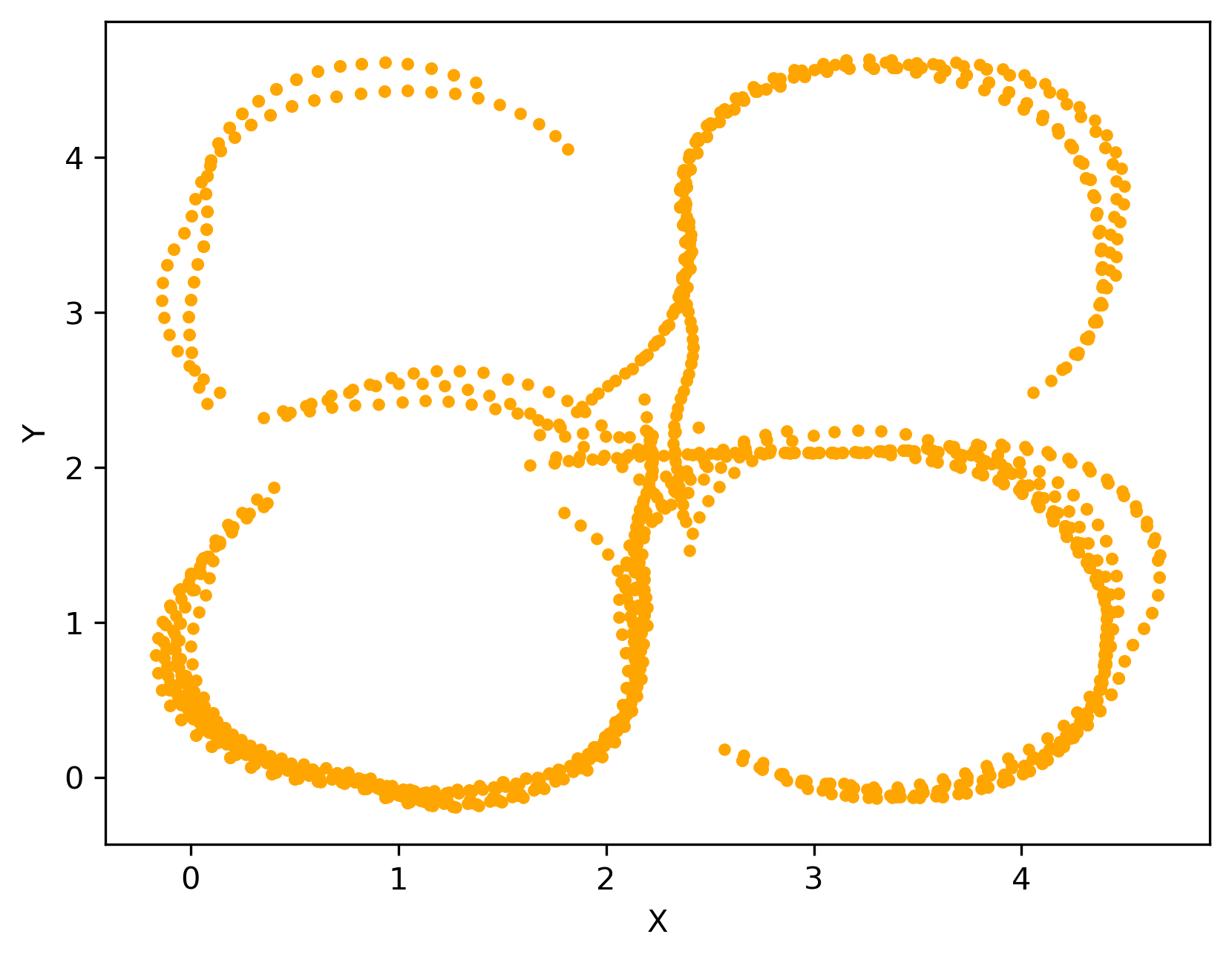}
    \caption{Lane Detection}
    \label{fig:Noise2}
\end{subfigure}
\begin{subfigure}[b]{0.49\columnwidth}
    \centering
    \includegraphics[width=0.99\columnwidth,trim={5px, 8px, 0, 0},clip]{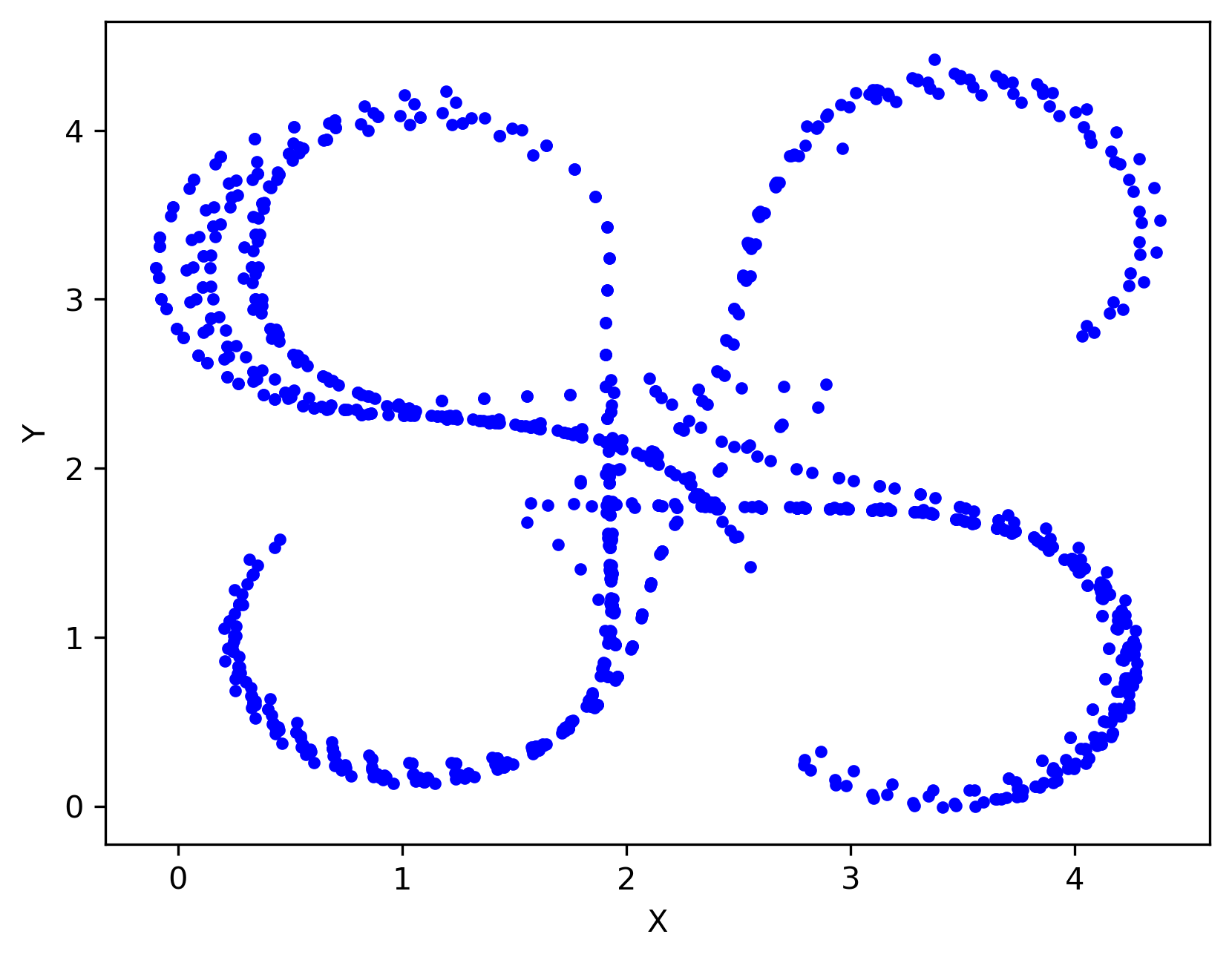}
    \caption{Speeding}
    \label{fig:Noise3}
\end{subfigure}
\begin{subfigure}[b]{0.49\columnwidth}
    \centering
    \includegraphics[width=0.99\columnwidth,trim={5px, 8px, 0, 0},clip]{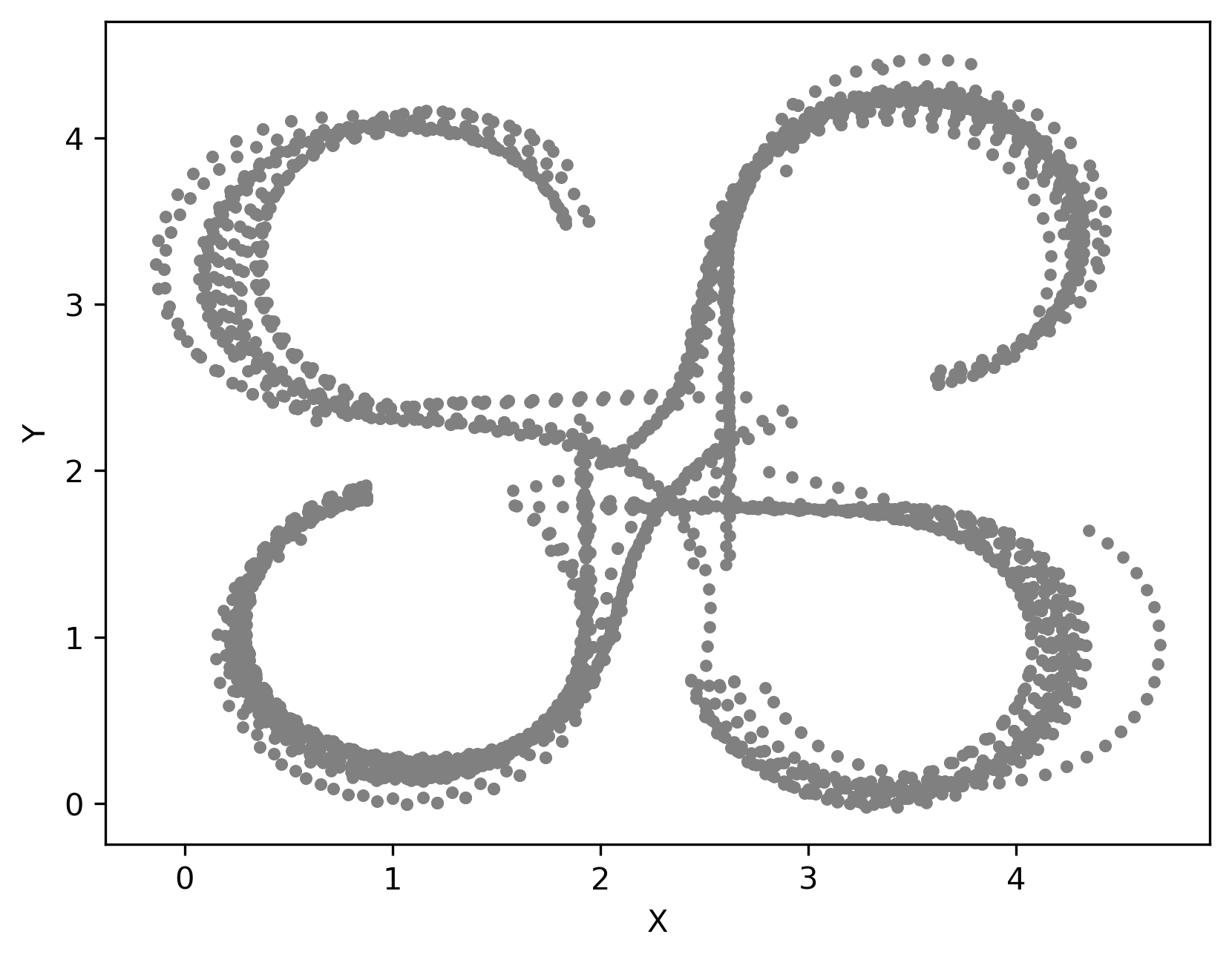}
    \caption{Nominal}
    \label{fig:Noiseless}
\end{subfigure}
\caption{Trajectories collected in the MiniCity used for training and evaluation.}
\label{fig:all_trajectories}
\end{figure}

During training, a single vehicle navigates in the MiniCity autonomously, with human monitoring and handovers in case of on or off-road collisions. Figure~\ref{fig:all_trajectories} shows the training poses captured for four failure modes: periodic noise, lane shift, speeding, and nominal driving. In addition, a manual joystick and first-person-view steering setup can be used to collect manual reckless driving. The ability to deploy autonomously with multiple vehicles enables large-scale collection of driving, for a total of over $3$ hours of driving data. Additionally, we augment the dataset of collected trajectories by applying a sliding window to generate additional sequences for training. 

To simulate the consequences of vehicle failures that may not manifest purely in driving style, we simultaneously deactivate the vehicle's collision avoidance and traffic light compliance. During training, we intentionally ignore these effects by training with a single vehicle (removing effects of collision avoidance) and removing trajectories immediately before and after the intersection (affected by the deactivated traffic light following) so as to not bias the model towards these effects. Similarly, during deployment, we do not evaluate predictions immediately before and after the intersection.  

\subsection{Detecting Failures and Warning Cross Traffic}

The intelligent traffic lights in the MiniCity monitor the oncoming traffic positions, both to provide reservation-based traffic management~\cite{Buckman2019} for nominal autonomous driving and to monitor the traffic for anomalies. In this work, we focus on learning to detect anomalous drivers directly from pose information, as such, the traffic manager accesses the pose information published from the motion capture. FailureNet receives each vehicle's pose at 2Hz and inputs a sequence of $L$ previous poses into the RNN. If fewer than $L$ poses have been received, the detector does not output a prediction. If FailureNet's output is above a detection threshold $\bar{Z}$, then a warning is sent to AVs approaching the intersection. Vehicles outside the intersection entrance do not receive warnings and can proceed normally.

\begin{table*}[ht]
    \centering
        \caption{FailureNet Accuracy on Validation Data}
        \begin{adjustbox}{width=0.7\textwidth}
    \begin{tabular}{lccccccc}
    \toprule
        & \# learnable & \multicolumn{4}{c}{\textbf{Accuracy in $\%$}}   \\
        \textbf{Method} &   parameters & All & Periodic & Lane Shift & Manual & Speeding & Nominal  \\
        \midrule
        Speed Threshold & 0 & 70.68 & 83.59 & 7.22 & 0.33 &  100.00  & 99.37 \\
        Speed + MLP & 5,569  & 74.86 & 97.95 & 37.91 & 32.67 & 100.00 & 88.10 \\
        Kalman Filter & 0 & 60.29 & 92.30 & 75.86 & 81.48 & 100.00 & 36.79 \\
        FFT  Threshold & 0 & 55.40  & 0.0 & 0.00 & 5.67 & 4.73 & 99.19 \\
        FFT + MLP & 6,209  & 93.00 & 95.38 & 92.06 & 73.67 & 100.00 & 97.11 \\  
        MLP & 8,129 &  97.44 & 96.41 & 97.11 & 93.67 & 100.00 & 98.38 \\        

\midrule
        FailureNet-LSTM & 26,049 & 98.42 & 97.44 & 98.19 & 96.33 & 99.32 & 99.10 \\
        FailureNet-GRU & 21,633 &  97.78 & 94.36 & 95.67 & 95.00 & 98.65 & 99.55 \\
        FailureNet-CfC & \textbf{1,936} & 97.78 & 92.82 & 100.00 & 92.33 & 98.65 & 99.46 \\
        \bottomrule
    \end{tabular}
    \end{adjustbox}
    \label{tab:accuracy_results}
\end{table*}

\section{RESULTS}

\subsection{Baselines}
\noindent We implement non-RNN baselines to benchmark the performance of the RNN failure estimators. The baselines include two thresholds based on filtering the input, a Kalman Filter, and a multi-layer perceptron (MLP).

\subsubsection{Speed Threshold}
The vehicle speed is computed based on previous $L$ poses and a threshold is computed based on either the average or max speed. We compute an estimate $\hat{z} =\frac{1}{L} \sum_i^L s_t \geq \bar{S}$ or $\hat{z}_{\max} = \max s_t$ where $s_t = ||x_t^2 + y_t^2||^{1/2}$. We iterate over possible thresholding values and choice of maximum or average speed, and choose a threshold that maximizes overall validation accuracy.

\subsubsection{Fast Fourier Transform (FFT) Power Threshold}

For noisy inputs, we first compute the FFT of the trajectories, to distinguish between noise profiles applied at the steering and speed. We compute the one-dimensional FFT of the vehicle yaw and select the higher-order modes for thresholding, $\omega_{2} \ldots \omega_{L/2}$. We choose a threshold on the maximum or average spectral power ($P(\omega_k) = |\omega_k|^2$) of the sequence, by searching through possible thresholds $\bar{P}$ which produces the highest validation accuracy.

\subsubsection{Kalman Filter} A Kalman filter is also applied to noisy vehicle trajectories to evaluate the failure of the trajectory. To be specific, the failure is evaluated by setting a threshold $\delta_{\kappa}$ to the measurement post-fit residual of the Kalman filter, which is absolute distance between the observation and filter predicted position. The Kalman state is in dimension 6, which includes 3-dimensional position/orientation and corresponding velocities. The optimal threshold $\delta_{\kappa}$ is 0.2, which was found by grid search.

\subsubsection{Multi-Layer Perceptron (MLP)}
We explore three different multi-layer perceptrons (MLP) with varying inputs to classify the failure state. We tune a standard MLP to determine the number of hidden layers, dimension, and dropout to maximize accuracy with the input being a concatenation of the poses in the previous $L$ timesteps. We train two additional networks, with the same MLP architecture, but add a pre-filter at the network input which computes either the speeds $s_t$ or FFT of the inputs $\omega_k$.

\subsection{Model Accuracy on Validation Data}
We evaluate the accuracy of our neural network based on the true positives (TP) and true negatives (TN), where \( \text{accuracy} = \frac{TP + TN}{P + N} \), where positive (P) samples correspond to vehicles with failure modes and negative (N) are nominal drivers. In both evaluation and deployment, a positive (failure) detection is threshold at $\hat{z} > 0.5$.

In Table~\ref{tab:accuracy_results}, we report the detection accuracy for each method on both the entire validation dataset (failure modes and nominal drivers), as well as accuracy in detecting each individual failure mode.  The RNN architectures (LSTM, GRU, CfC) provide the highest accuracy rates over the baselines. FailureNet-LSTM is overall the best performing, with highest accuracy on the most difficult failure mode (manual driver). FailureNet-GRU and FailureNet-CfC provide the highest true negative rate on the nominal driver validation data. One advantage of FailureNet-CfC is its relatively small size compared to the other RNNs and MLP (which require $10\times$ and $4\times$ the parameters, respectively).

\begin{figure}[htb]
\begin{subfigure}[c]{0.99\columnwidth}
    \centering
    \includegraphics[width=0.49\columnwidth,trim={300px 200px 300px 50px},clip]{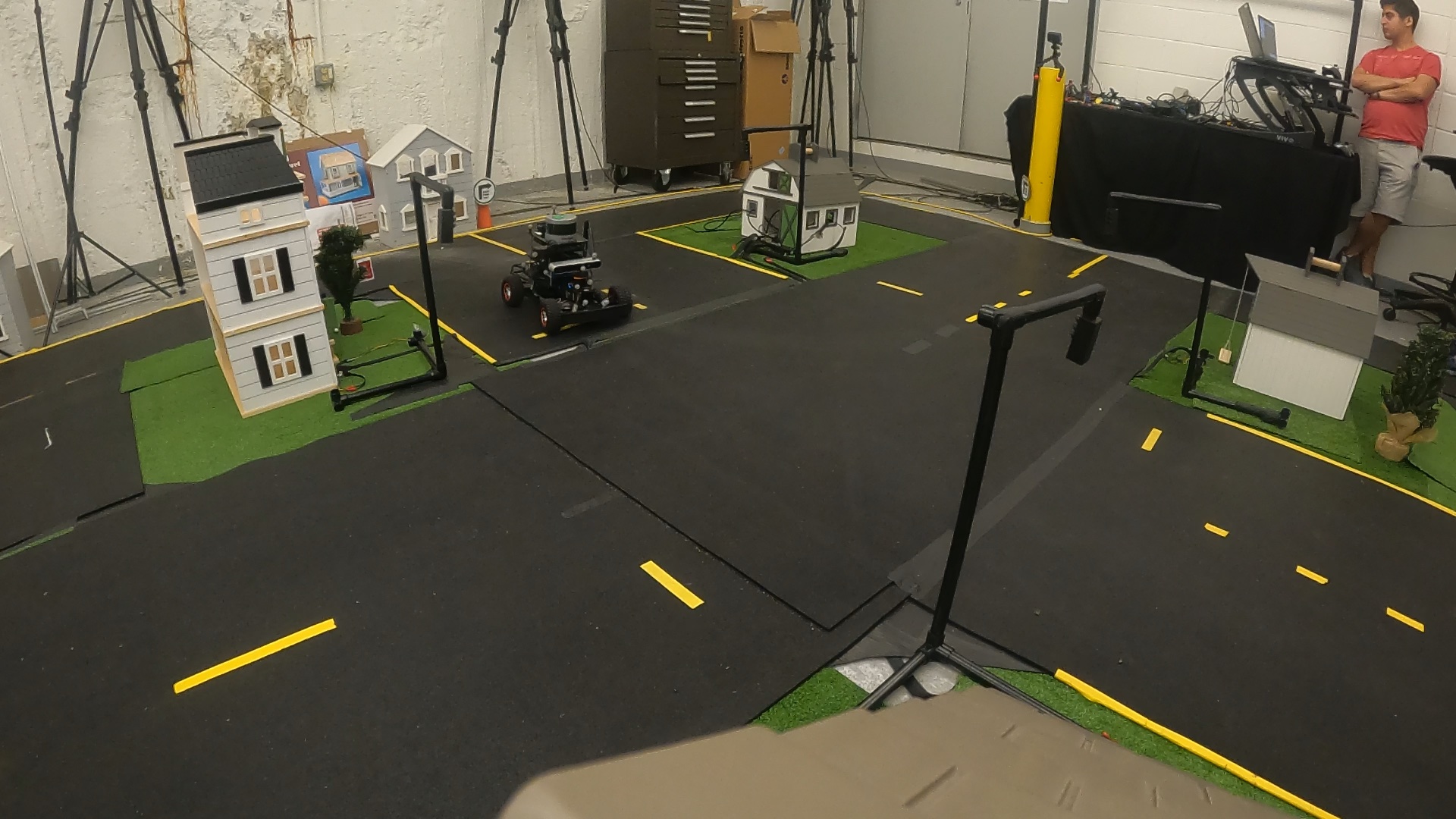}     
    \hfill
    \centering
    \includegraphics[width=0.49\columnwidth, trim={0px 200px 0 0}, clip]{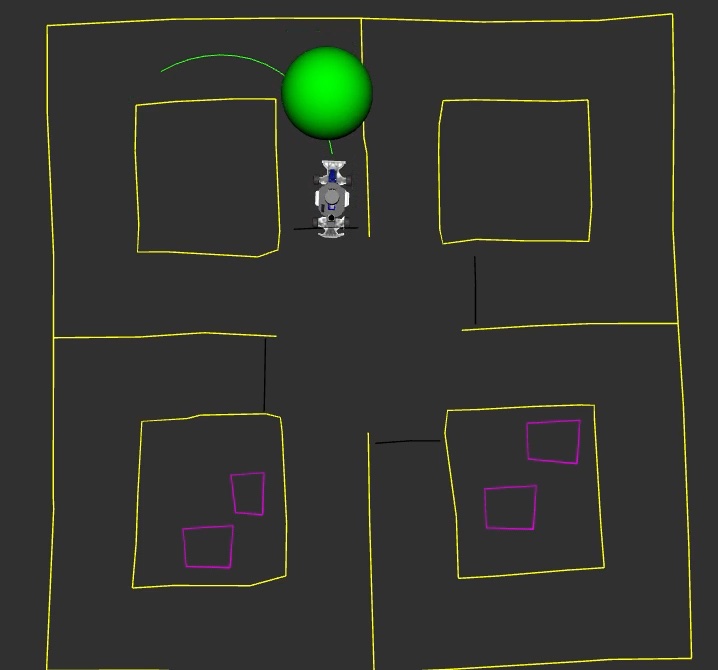}    
    \caption{Nominal Driving}
\end{subfigure}
\vskip\baselineskip

\begin{subfigure}[c]{0.99\columnwidth}
\centering
     \includegraphics[width=0.49\columnwidth,trim={300px 200px 300px 50px},clip]{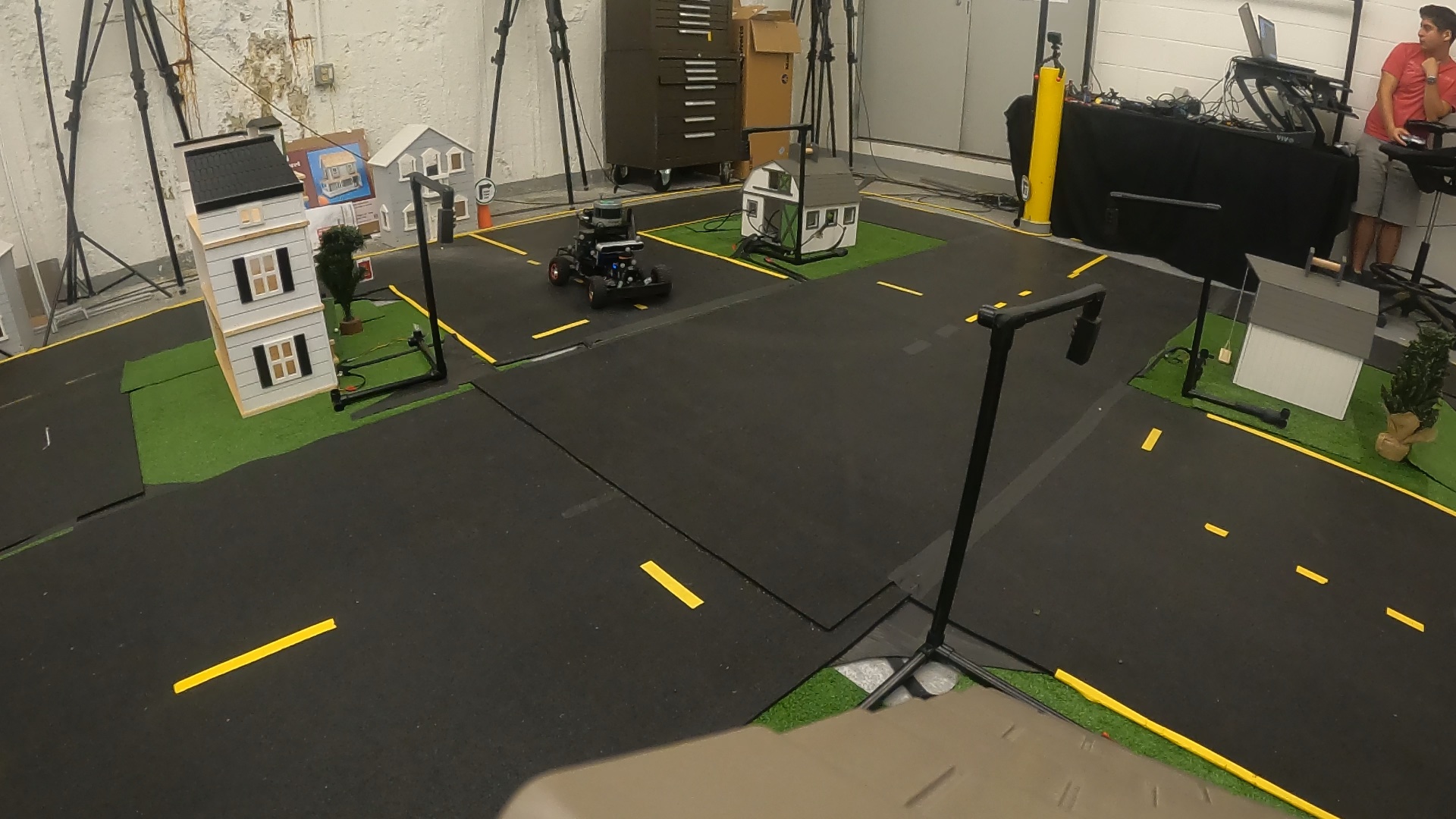}
     \hfill
    \includegraphics[width=0.49\columnwidth, trim={0px 200px 0 0}, clip]{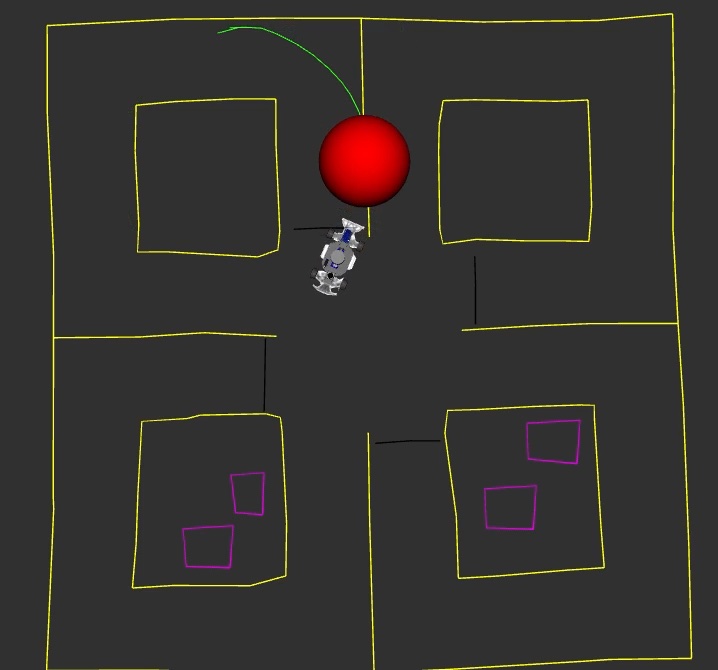}     
    \caption{Lane Detector Failure}
    
\end{subfigure}
    \caption{FailureNet deployed in the MiniCity distinguishes between nominal drivers and reckless drivers. Input sequence of poses (green line) are used by the network to output a prediction of the vehicle's failure status (red/green sphere).}
    \label{fig:example:reckless}
\end{figure}
\subsection{Safety Evaluation in the MiniCity}
Finally, we deploy FailureNet in the MiniCity with two vehicles, one that drives nominally and one that drives with one of the failure mode activated.  Figure~\ref{fig:example:reckless} shows the detector deployed in the MiniCity, with the input sequence and prediction visualized. We deploy each method and failure mode for 3 minutes each and evaluate the accuracy of the detector, running at 1Hz. In Table~\ref{tab:hardware_results}, we report the accuracy for various baselines and RNNs, in various failure settings.  When evaluating FailureNet, if a manual handover is required (such as immediately before or after a collision), then we do not record detections immediately before and after a collision. This ensures that we do not unintentionally reward FailureNet for identifying manual takeover maneuvers. We find that FailureNet-LSTM and Failurenet-CfC perform best in the MiniCity, with an overall accuracy of 84\%. The speed threshold performs well on the failure modes with speed components, however, fails to identify the lane shift failure mode since speed is unaffected. In contrast, our approach performs well across all failure modes and outperforms the MLP when evaluated online.  

\begin{table}[t]
    \centering
    \caption{FailureNet Accuracy [\%] Deployed in MiniCity }
    \begin{adjustbox}{width=0.99\columnwidth}
    \begin{tabular}{lcccccc}
    \toprule
        \textbf{Method} & All & Periodic & Lane Shift & Manual & Speeding & Nominal  \\
        \midrule
Speed Threshold &  73 &  100 &  8 &  98 &  100 &  100  \\ 
FFT Threshold &  21 &  13 &  1 &  7 &  2 &  98  \\ 
Kalman Filter & 71 &  75 &  78 &  92 &  94 &  21 \\
MLP &  74 &  65 &  75 &  79 &  64 &  88  \\ 
\midrule
FailureNet-LSTM &  \textbf{84} &  79 &  90 &  95 &  69 &  84  \\ 
FailureNet-GRU &  79 &  56 &  88 &  86 &  64 &  95  \\ 
FailureNet-CfC & \textbf{84} &  79 &  87 &  78 &  85 &  87  \\
        \bottomrule
    \end{tabular}
    \end{adjustbox}
    \label{tab:hardware_results}
\end{table}

\section{CONCLUSIONS}

In this work, we present an end-to-end method for identifying failure modes presented by drivers as they approach an intersection. 
We utilize a 1/10th scale MiniCity to generate a dataset of various driving and vehicle failure modes to train an RNN to identify drivers with failures in planning and control. 
When deployed in the MiniCity, FailureNet accurately detects vehicle failure and can provide proactive warnings to oncoming traffic. 
While we train and validate our approach on only a subset of failure modes, our approach is general to various types of failure modes.
In addition, the use of a miniature platform for learning to detect rare and dangerous events can be extended to future studies.

\addtolength{\textheight}{-1cm}   


\clearpage
\bibliographystyle{IEEEtran}
\bibliography{IEEEabrv,references}

\begin{thebibliography}{10}
\providecommand{\url}[1]{#1}
\csname url@rmstyle\endcsname
\providecommand{\newblock}{\relax}
\providecommand{\bibinfo}[2]{#2}
\providecommand\BIBentrySTDinterwordspacing{\spaceskip=0pt\relax}
\providecommand\BIBentryALTinterwordstretchfactor{4}
\providecommand\BIBentryALTinterwordspacing{\spaceskip=\fontdimen2\font plus
\BIBentryALTinterwordstretchfactor\fontdimen3\font minus
  \fontdimen4\font\relax}
\providecommand\BIBforeignlanguage[2]{{%
\expandafter\ifx\csname l@#1\endcsname\relax
\typeout{** WARNING: IEEEtran.bst: No hyphenation pattern has been}%
\typeout{** loaded for the language `#1'. Using the pattern for}%
\typeout{** the default language instead.}%
\else
\language=\csname l@#1\endcsname
\fi
#2}}

\bibitem{Hecker2018}
S.~Hecker, D.~Dai, and L.~{Van Gool}, ``{Failure Prediction for Autonomous
  Driving},'' \emph{IEEE Intelligent Vehicles Symposium, Proceedings}, vol.
  2018-June, no.~Iv, pp. 1792--1799, 2018.

\bibitem{Svegliato2019}
J.~Svegliato, K.~H. Wray, S.~J. Witwicki, J.~Biswas, and S.~Zilberstein,
  ``{Belief Space Metareasoning for Exception Recovery},'' in \emph{2019
  IEEE/RSJ International Conference on Intelligent Robots and Systems
  (IROS)}.\hskip 1em plus 0.5em minus 0.4em\relax IEEE, nov 2019, pp.
  1224--1229.

\bibitem{Zhang2017a}
M.~Zhang, C.~Chen, T.~Wo, T.~Xie, M.~Z.~A. Bhuiyan, and X.~Lin, ``{SafeDrive:
  Online Driving Anomaly Detection From Large-Scale Vehicle Data},'' \emph{IEEE
  Transactions on Industrial Informatics}, vol.~13, no.~4, pp. 2087--2096,
  2017.

\bibitem{Johnson2011b}
D.~A. Johnson and M.~M. Trivedi, ``{Driving style recognition using a
  smartphone as a sensor platform},'' \emph{IEEE Conference on Intelligent
  Transportation Systems, Proceedings, ITSC}, pp. 1609--1615, 2011.

\bibitem{Vasconcelos2017}
I.~Vasconcelos, R.~O. Vasconcelos, B.~Olivieri, M.~Roriz, M.~Endler, and M.~C.
  Junior, ``{Smartphone-based outlier detection: a complex event processing
  approach for driving behavior detection},'' \emph{Journal of Internet
  Services and Applications}, vol.~8, no.~1, 2017.

\bibitem{Quintero2010}
G.~C.~M. Quintero, J.~A.~O. Lopez, and J.~M.~P. Rua, ``{Intelligent erratic
  driving diagnosis based on artificial neural networks},'' in \emph{2010 IEEE
  ANDESCON}.\hskip 1em plus 0.5em minus 0.4em\relax IEEE, sep 2010, pp. 1--6.

\bibitem{Siddiqui2016uai}
A.~Siddiqui, A.~Fern, T.~G. Dietterich, and S.~Das, ``{Finite Sample Complexity
  of Rare Pattern Anomaly Detection},'' in \emph{UAI'16: Proceedings of the
  Thirty-Second Conference on Uncertainty in Artificial Intelligence}, 2016,
  pp. 686--695.

\bibitem{Wu2021arxiv}
T.~Wu and J.~Ortiz, ``{RLAD: Time Series Anomaly Detection through
  Reinforcement Learning and Active Learning},'' Mar 2021.

\bibitem{Ryan2021}
C.~Ryan, F.~Murphy, and M.~Mullins, ``{End-to-End Autonomous Driving Risk
  Analysis: A Behavioural Anomaly Detection Approach},'' \emph{IEEE
  Transactions on Intelligent Transportation Systems}, vol.~22, no.~3, pp.
  1650--1662, 2021.

\bibitem{Kawashima2003}
M.~Hashimoto, H.~Kawashima, and F.~Oba, ``A multi-model based fault detection
  and diagnosis of internal sensors for mobile robot,'' in \emph{Proceedings
  2003 IEEE/RSJ International Conference on Intelligent Robots and Systems
  (IROS 2003)}, vol.~4, 2003, pp. 3787--3792 vol.3.

\bibitem{Duan2005}
Z.~H. Duan, Z.~X. Cai, and J.~X. Yu, ``{Fault diagnosis and fault tolerant
  control for wheeled mobile robots under unknown environments: A survey},''
  \emph{Proceedings - IEEE International Conference on Robotics and
  Automation}, vol. 2005, no. April, pp. 3428--3433, 2005.

\bibitem{Takei2005}
Y.~Takei and Y.~Furukawa, ``{Estimate of driver's fatigue through steering
  motion},'' \emph{Conference Proceedings - IEEE International Conference on
  Systems, Man and Cybernetics}, vol.~2, no.~1, pp. 1765--1770, 2005.

\bibitem{DiBiase2021}
G.~{Di Biase}, H.~Blum, R.~Siegwart, and C.~Cadena, ``{Pixel-wise Anomaly
  Detection in Complex Driving Scenes},'' \emph{Proceedings of the IEEE
  Computer Society Conference on Computer Vision and Pattern Recognition}, pp.
  16\,913--16\,922, 2021.

\bibitem{Doshi2009}
A.~Doshi and M.~M. Trivedi, ``{On the roles of eye gaze and head dynamics in
  predicting driver's intent to change lanes},'' \emph{IEEE Transactions on
  Intelligent Transportation Systems}, vol.~10, no.~3, pp. 453--462, 2009.

\bibitem{Fletcher2009}
L.~Fletcher and A.~Zelinsky, ``{Driver Inattention Detection based on Eye
  Gaze–Road Event Correlation},'' \emph{International Journal of Robotics
  Research}, vol.~28, no.~6, pp. 774--801, 2009.

\bibitem{wiederer2022}
J.~Wiederer, A.~Bouazizi, M.~Troina, U.~Kressel, and V.~Belagiannis, ``Anomaly
  detection in multi-agent trajectories for automated driving,'' in
  \emph{Proceedings of the 5th Conference on Robot Learning}, ser. Proceedings
  of Machine Learning Research, A.~Faust, D.~Hsu, and G.~Neumann, Eds., vol.
  164.\hskip 1em plus 0.5em minus 0.4em\relax PMLR, 08--11 Nov 2022, pp.
  1223--1233.

\bibitem{Morton2017}
J.~Morton, T.~A. Wheeler, and M.~J. Kochenderfer, ``{Analysis of Recurrent
  Neural Networks for Probabilistic Modeling of Driver Behavior},'' \emph{IEEE
  Transactions on Intelligent Transportation Systems}, vol.~18, no.~5, pp.
  1289--1298, 2017.

\bibitem{Salzmann2020}
T.~Salzmann, B.~Ivanovic, P.~Chakravarty, and M.~Pavone, ``Trajectron++:
  Dynamically-feasible trajectory forecasting with heterogeneous data,'' in
  \emph{Computer Vision -- ECCV 2020}, A.~Vedaldi, H.~Bischof, T.~Brox, and
  J.-M. Frahm, Eds.\hskip 1em plus 0.5em minus 0.4em\relax Cham: Springer
  International Publishing, 2020, pp. 683--700.

\bibitem{Shirazi2017}
M.~S. Shirazi and B.~T. Morris, ``{Looking at Intersections: A Survey of
  Intersection Monitoring, Behavior and Safety Analysis of Recent Studies},''
  \emph{IEEE Transactions on Intelligent Transportation Systems}, vol.~18,
  no.~1, pp. 4--24, 2017.

\bibitem{Bjorklund2005}
G.~M. Bj{\"{o}}rklund and L.~{\AA}berg, ``{Driver behaviour in intersections:
  Formal and informal traffic rules},'' \emph{Transportation Research Part F:
  Traffic Psychology and Behaviour}, vol.~8, no.~3, pp. 239--253, 2005.

\bibitem{Sun2022}
J.~Sun, S.~Kousik, D.~Fridovich-Keil, and M.~Schwager, ``{Self-Supervised
  Traffic Advisors: Distributed, Multi-view Traffic Prediction for Smart
  Cities},'' \emph{arXiv preprint}, 2022.

\bibitem{Phillips2017}
D.~J. Phillips, T.~A. Wheeler, and M.~J. Kochenderfer, ``{Generalizable
  intention prediction of human drivers at intersections},'' \emph{IEEE
  Intelligent Vehicles Symposium, Proceedings}, no.~Iv, pp. 1665--1670, 2017.

\bibitem{Lefevre2012}
S.~Lef{\`{e}}vre, C.~Laugier, and J.~Iba{\~{n}}ez-Guzm{\'{a}}n, ``{Risk
  assessment at road intersections: Comparing intention and expectation},''
  \emph{IEEE Intelligent Vehicles Symposium, Proceedings}, pp. 165--171, 2012.

\bibitem{raja2022ai}
G.~Raja, M.~Begum, S.~Gurumoorthy, D.~S. Rajendran, P.~Srividya, K.~Dev, and
  N.~M.~F. Qureshi, ``Ai-empowered trajectory anomaly detection and
  classification in 6g-v2x,'' \emph{IEEE Transactions on Intelligent
  Transportation Systems}, 2022.

\bibitem{Salim2007}
F.~D. Salim, S.~W. Loke, A.~Rakotonirainy, B.~Srinivasan, and S.~Krishnaswamy,
  ``{Collision pattern modeling and Real-Time collision detection at road
  intersections},'' \emph{IEEE Conference on Intelligent Transportation
  Systems, Proceedings, ITSC}, pp. 161--166, 2007.

\bibitem{Kowshik2011a}
H.~Kowshik, D.~Caveney, and P.~R. Kumar, ``{Provable systemwide safety in
  intelligent intersections},'' \emph{IEEE Transactions on Vehicular
  Technology}, vol.~60, no.~3, pp. 804--818, 2011.

\bibitem{Dresner2008a}
K.~Dresner and P.~Stone, ``{Mitigating catastrophic failure at intersections of
  autonomous vehicles},'' \emph{Proceedings of the International Joint
  Conference on Autonomous Agents and Multiagent Systems, AAMAS}, vol.~3, pp.
  1361--1364, 2008.

\bibitem{Yu2019}
B.~Yu, S.~Bao, F.~Feng, and J.~Sayer, ``{Examination and prediction of drivers'
  reaction when provided with V2I communication-based intersection maneuver
  strategies},'' \emph{Transportation Research Part C: Emerging Technologies},
  vol. 106, no. November 2018, pp. 17--28, 2019.

\bibitem{Feng2018}
Y.~Feng, C.~Yu, S.~Xu, H.~X. Liu, and H.~Peng, ``{An Augmented Reality
  Environment for Connected and Automated Vehicle Testing and Evaluation},''
  \emph{IEEE Intelligent Vehicles Symposium, Proceedings}, vol. 2018-June,
  no.~Iv, pp. 1549--1554, 2018.

\bibitem{NationalCenterforStatisticsandAnalysis2022}
{National Center for Statistics and Analysis}, ``Speeding: 2020 data,''
  National Highway Traffic Safety Administration, Tech. Rep. June, 2022,
  {Traffic Safety Facts. Report No. DOT HS 813 320}.

\bibitem{hochreiter1997long}
S.~Hochreiter and J.~Schmidhuber, ``Long short-term memory,'' \emph{Neural
  computation}, vol.~9, no.~8, pp. 1735--1780, 1997.

\bibitem{Chung2014}
J.~Chung, C.~Gulcehre, K.~Cho, and Y.~Bengio, ``\BIBforeignlanguage{English
  (US)}{Empirical evaluation of gated recurrent neural networks on sequence
  modeling},'' in \emph{\BIBforeignlanguage{English (US)}{NIPS 2014 Workshop on
  Deep Learning, December 2014}}, 2014.

\bibitem{lechner2019designing}
M.~Lechner, R.~Hasani, M.~Zimmer, T.~A. Henzinger, and R.~Grosu, ``Designing
  worm-inspired neural networks for interpretable robotic control,'' in
  \emph{2019 International Conference on Robotics and Automation (ICRA)}.\hskip
  1em plus 0.5em minus 0.4em\relax IEEE, 2019, pp. 87--94.

\bibitem{lechner2020neural}
M.~Lechner, R.~Hasani, A.~Amini, T.~A. Henzinger, D.~Rus, and R.~Grosu,
  ``Neural circuit policies enabling auditable autonomy,'' \emph{Nature Machine
  Intelligence}, vol.~2, no.~10, pp. 642--652, 2020.

\bibitem{vorbach2021causal}
C.~Vorbach, R.~Hasani, A.~Amini, M.~Lechner, and D.~Rus, ``Causal navigation by
  continuous-time neural networks,'' \emph{Advances in Neural Information
  Processing Systems}, vol.~34, 2021.

\bibitem{Hasani2021liquid}
R.~Hasani, M.~Lechner, A.~Amini, D.~Rus, and R.~Grosu, ``Liquid time-constant
  networks,'' \emph{Proceedings of the AAAI Conference on Artificial
  Intelligence}, vol.~35, no.~9, pp. 7657--7666, May 2021.

\bibitem{gu2022efficiently}
A.~Gu, K.~Goel, and C.~Re, ``Efficiently modeling long sequences with
  structured state spaces,'' in \emph{International Conference on Learning
  Representations}, 2022.

\bibitem{chen2018neural}
T.~Q. Chen, Y.~Rubanova, J.~Bettencourt, and D.~K. Duvenaud, ``Neural ordinary
  differential equations,'' in \emph{Advances in neural information processing
  systems}, 2018, pp. 6571--6583.

\bibitem{hasani2021closed}
R.~Hasani, M.~Lechner, A.~Amini, L.~Liebenwein, M.~Tschaikowski, G.~Teschl, and
  D.~Rus, ``Closed-form continuous-depth models,'' \emph{arXiv preprint
  arXiv:2106.13898}, 2021.

\bibitem{friston2003dynamic}
K.~J. Friston, L.~Harrison, and W.~Penny, ``Dynamic causal modelling,''
  \emph{Neuroimage}, vol.~19, no.~4, pp. 1273--1302, 2003.

\bibitem{Buckman2022sensors}
N.~Buckman, A.~Hansen, S.~Karaman, and D.~Rus, ``{Evaluating Autonomous Urban
  Perception and Planning in a 1/10th Scale MiniCity},'' \emph{Sensors},
  vol.~22, no.~18, p. 6793, sep 2022.

\bibitem{Buckman2019}
N.~Buckman, A.~Pierson, W.~Schwarting, S.~Karaman, and D.~Rus, ``{Sharing is
  Caring: Socially-Compliant Autonomous Intersection Negotiation},'' in
  \emph{2019 IEEE/RSJ International Conference on Intelligent Robots and
  Systems (IROS)}.\hskip 1em plus 0.5em minus 0.4em\relax IEEE, nov 2019, pp.
  6136--6143.

\end{thebibliography}

\end{document}